\documentclass[sigconf]{acmart}

\usepackage{textcomp}
\usepackage{tabularx}
\usepackage{makecell}
\usepackage{booktabs}
\usepackage{subfig}
\usepackage{relsize}
\usepackage{placeins}
\usepackage{seqsplit}

\renewcommand\footnotetextcopyrightpermission[1]{}
\copyrightyear{}
\acmYear{}
\setcctype{}
\acmConference[]{}{}{}
\acmDOI{}
\acmISBN{}

\acmSubmissionID{147}

\begin{document}


\title{SweetSpot: An Analytical Model for Predicting Energy Efficiency of LLM Inference}


\author{Hiari Pizzini Cavagna}
\authornote{These authors contributed equally to this work.}
\email{hiari.pizzinicavagna@unibo.it}
\orcid{0009-0005-2768-0418}
\affiliation{%
  \institution{University of Bologna}
  \city{Bologna}
  \state{BO}
  \country{Italy}
}

\author{Andrea Proia}
\authornotemark[1]
\email{andrea.proia@unibo.it}
\orcid{0009-0005-6367-762X}
\affiliation{%
  \institution{University of Bologna}
  \city{Bologna}
  \state{BO}
  \country{Italy}
}

\author{Giacomo Madella}
\email{giacomo.madella@unibo.it}
\orcid{0009-0009-1289-6354}
\affiliation{%
  \institution{University of Bologna}
  \city{Bologna}
  \state{BO}
  \country{Italy}
}

\author{Giovanni B. Esposito}
\email{g.esposito@unibo.it}
\orcid{0009-0002-9595-2962}
\affiliation{%
  \institution{University of Bologna}
  \city{Bologna}
  \state{BO}
  \country{Italy}
}

\author{Francesco Antici}
\email{francesco.antici@unibo.it}
\orcid{0000-0002-1125-0588}
\affiliation{%
  \institution{University of Bologna}
  \city{Bologna}
  \state{BO}
  \country{Italy}
}

\author{Daniele Cesarini}
\email{d.cesarini@cineca.it}
\orcid{0000-0003-1294-372X}
\affiliation{%
  \institution{Cineca}
  \city{Casalecchio di Reno}
  \state{BO}
  \country{Italy}
}

\author{Zeynep Kiziltan}
\email{zeynep.kiziltan@unibo.it}
\orcid{0000-0003-0412-4396}
\affiliation{%
  \institution{University of Bologna}
  \city{Bologna}
  \state{BO}
  \country{Italy}
}

\author{Andrea Bartolini}
\email{a.bartolini@unibo.it}
\orcid{0000-0002-1148-2450}
\affiliation{%
  \institution{University of Bologna}
  \city{Bologna}
  \state{BO}
  \country{Italy}
}

\renewcommand{\shortauthors}{H. Pizzini Cavagna, A. Proia, et al.}

\begin{abstract}
    Large Language Models (LLMs) inference is central to modern AI applications, dominating worldwide datacenter workloads, making it critical to predict its energy footprint. 
    Existing approaches estimate energy consumption as a simple linear function of input and output sequence. However, by analyzing the autoregressive structure of Transformers, which implies a fundamentally non-linear relationship between input and output sequence lengths and energy consumption, we demonstrate the existence of a generation energy minima. 
    Peak efficiency occurs with short-to-moderate inputs and medium-length outputs, while efficiency drops sharply for long inputs or very short outputs.
    Consequently, we propose \textit{SweetSpot}, an analytical model derived from the computational and memory-access complexity of the Transformer architecture, which accurately characterizes the efficiency curve as a function of input and output lengths.
    To assess accuracy, we measure energy consumption using TensorRT-LLM on NVIDIA H100 GPUs across a diverse set of LLMs ranging from 1B to 9B parameters, including OPT, LLaMA, Gemma, Falcon, Qwen2, and Granite. We test input and output lengths from 64 to 4096 tokens and achieve a mean MAPE of 1.79\%.
    Our results show that aligning sequence lengths with these efficiency “sweet spots” reduce energy usage, up to $33.41\times$, enabling informed truncation, summarization, and adaptive generation strategies in production systems.
\end{abstract}





\maketitle
\section{Introduction}
Large Language Models (LLMs) have rapidly emerged as the foundation of modern Natural Language Processing (NLP), enabling applications ranging from reasoning and summarization to dialogue systems and generative Artificial Intelligence (AI). Their adoption at scale, however, has introduced a growing concern: the substantial energy consumption associated with inference, which has been estimated to account for up to 90\% of the model’s life-cycle energy consumption \cite{jegham2025hungryaibenchmarkingenergy}. Unlike training, which is an episodic process, inference occurs continuously in production settings, every time an LLM generates output for a user query. This shift makes inference efficiency a critical challenge for both sustainability and deployment cost.  

A key contributor to inference complexity in LLMs is the attention mechanism.
Since the original Transformer design \cite{vaswani2017}, the standard Multi-Head Attention (MHA) has undergone several modifications aimed at improving efficiency. 
Variants such as Multi-Query Attention (MQA) \cite{shazer2019MQA} and Grouped-Query Attention (GQA) \cite{ainslie2023GQA} reduce memory requirements by sharing Key–Value (KV) projections across heads, thereby lowering the cost of maintaining and accessing the KV cache during decoding. 
Recent studies \cite{dao2022flashattention} further highlight that attention kernels are often memory-bound rather than compute-bound, making their design a critical factor for inference efficiency at scale. 
These insights underscore the central role of attention in determining both the computational and energy footprint of LLM inference.

The energy consumption of LLM inference is shaped by a complex interplay of model size, architecture, and workload geometry (i.e., the distribution of input and output sequence lengths), and its characterization requires thorough empirical validation.\cite{fernandez2025energyconsiderationslargelanguage} Recent studies on inference energy consumption focus only on specific models \cite{stojkovic2024greenerllmsbringingenergyefficiency} or coarse-grained metrics (e.g., latency and throughput \cite{maliakel2025investigatingenergyefficiencyperformance}), which fail to generalize and provide fine-grained inference energy profiling. Moreover, previous claims \cite{wilkins2024offlineenergyoptimalllmserving} on linear relation between the workload parameters ($n_{\text{in}}$ = input tokens, $n_{\text{out}}$ = output tokens) leads to the conclusion that every generated token incurs a constant computational and energy cost, as reflected by commercial AI providers that charge inference services based on the number of generated output tokens or a linear combination of input and output tokens. 
However, transformer structure and autoregressive generation process, with distinct prefill and decode phases, suggest a more complex dependency of computational cost, and consequently energy consumption, on both input and output sequence lengths. This complexity is not fully captured by current state-of-the-art analytical models.

To address these limitations, we conduct a thorough and systematic analysis of the energy consumption of a broad set of LLMs (from 1B to 9B parameters), and we explore efficiency across a wide range of input and output lengths (from 64 to 4096 tokens).
Following the results of our investigation, in this paper we present the following contributions: 
\begin{itemize}
    \item The formulation of \textit{SweetSpot}, an analytical model derived from computational and memory-access complexity of inference,
    capable of predicting the peak energy efficiency input and output lengths configuration. The proposed model reveals how quadratic input costs and linear decoding costs combine to produce the efficiency patterns observed in practice.
    \item We validate \textit{SweetSpot} through extensive measurements \allowbreak across multiple LLM families, showing that it accurately captures the location of peak energy efficiency configurations (“sweet spots”) as a function of input and output lengths, achieving a mean MAPE of 1.79\% ($\pm$ 0.61).
    \item We demonstrate that efficiency is not monotonically dependent on input or output sequence lengths. Instead, it exhibits distinct regimes: peak efficiency occurs for short-to-medium inputs combined with medium outputs, while efficiency sharply declines for long inputs or very short outputs, with peak energy efficiency "sweet-spots" exceeding the worst-case setting by a factor of 33.41$\times$. 
\end{itemize}
We conduct our study using TensorRT-LLM,\footnote{\url{https://github.com/NVIDIA/TensorRT-LLM}} a state-of-the-art inference stack optimized for modern GPUs, running on NVIDIA H100 accelerators. Through TensorRT-LLM, we test several widely-used LLM models, such as OPT \cite{opt_models}, LLaMA \cite{llama_models}, Gemma \cite{gemma_models}, Falcon \cite{falcon_models}, Qwen2 \cite{qwen_models}, and Granite \cite{granite_models}. Doing so allows us to obtain reliable and realistic results which apply to a broad range of different models and architectures.

Our results suggest that production deployments can optimize energy efficiency by tailoring batching strategies, truncating overly long inputs, or aligning output lengths with the peak energy efficiency configurations predicted by our model.  

The rest of the paper is organized as follows. After giving the necessary background in Section~\ref{sec:background}, we review related work in Section~\ref{sec:related_works}. Then, in Section~\ref{sec:computation_model}, we present our analytical model \textit{SweetSpot}. In Section~\ref{sec:experimental_study} we explain our experimental methodology and illustrate the experimental results. Finally, we discuss our current limitations and future works in Section \ref{sec:limitations_future_works}, and conclude in Section \ref{sec:conclusions}.

\section{Background}
This section provides the necessary background for the analysis presented in this work. We briefly review the fundamentals of LLM architectures, their computational characteristics, and the main design principles of modern inference frameworks.
\label{sec:background}
\subsection{LLM Architecture}
LLMs are based on the Transformer architecture \cite{vaswani2017}, which has become the standard for generative AI. A transformer block is composed of two main components: the Multi-Head self-Attention mechanism (MHA) and the Feed-Forward Network (FFN), both surrounded by residual connections, layer normalization, and dropout. The attention mechanism is commonly implemented as MHA, where the hidden representation is projected into different subspaces (heads). Each head learns to capture different types of dependencies in the sequence, and their outputs are later combined. While this improves the model’s expressivity and ability to represent diverse relationships, it also increases the total number of operations roughly in proportion to the number of heads. To reduce this overhead, Multi-Query Attention (MQA) was proposed in 2019 \cite{shazer2019MQA}, modifying MHA by keeping multiple query heads but sharing a single set of Key-Value (KV) projections across all heads. This design drastically reduces memory usage and KV cache size during decoding, making inference more efficient. More recently, Grouped-Query Attention (GQA) was introduced in 2023 \cite{ainslie2023GQA} as a generalization of MHA and MQA: keys and values are shared within groups of heads, striking a balance between the efficiency of MQA and the expressivity of MHA. 

\subsection{LLM Inference Computation}
In large-scale models, the majority of parameters reside in the FFN layers, while the attention mechanism dominates runtime complexity due to its quadratic dependence on the input sequence length. 

LLM inference is typically divided into two phases: \textit{prefill}, where the full input sequence is processed in parallel to initialize the KV cache; and \textit{decode}, where tokens are generated autoregressively \cite{llminfrence_unveiled}. 
In the prefill stage, the computational cost is dominated by the quadratic attention term, whereas in the decode stage, the cost per token is linear with respect to the sequence length, due to the reuse of cached keys and values. Despite this, the decode phase often dominates total inference time for long outputs, as each token must be produced sequentially, limiting opportunities for parallelization.

On modern GPUs, the throughput of LLM inference is not solely limited by arithmetic performance but also by memory bandwidth. Storing and retrieving the large KV cache during decoding creates significant pressure on the memory subsystem with high bandwidth requirements. For instance, in autoregressive decoding, each generated token requires the loading of all previous keys and values from memory, resulting in attention kernels being strongly memory-bound, particularly for long contexts.

\subsection{Inference Frameworks}
When executing LLM inference on production hardware, several software for inference run-time exist, such as TensorRT-LLM \cite{tensorrtllm}, vLLM \cite{vllm} and DeepSpeed \cite{deepspeed}, each offering different levels of optimization and targeting different deployment constraints. They vary substantially in how they manage memory, parallelize attention, and schedule kernels. TensorRT-LLM focuses on graph-level optimizations and low-level kernel fusion, while vLLM is an open-source inference system explicitly tailored for high-throughput LLM serving. Its main contribution, the PagedAttention mechanism, improves memory efficiency by partitioning KV caches, thereby enabling scalable batch inference with reduced memory overhead. DeepSpeed, developed by Microsoft, provides a distributed training and inference framework aimed at maximizing scalability using techniques such as the Zero Redundancy Optimizer \cite{rajbhandari2020zeromemoryoptimizationstraining}.

\subsubsection{\textbf{TensorRT-LLM}}
In particular, TensorRT-LLM is NVIDIA’s specialized inference framework designed for optimizing LLM serving on modern GPUs. It builds upon the TensorRT optimization stack, extending it with kernels and execution strategies tailored for Transformer-based architectures. The framework provides a toolchain that converts pretrained models into highly optimized TensorRT engines. This process includes parsing the original model graph, applying operator fusion, and lowering precision to FP16, BF16, or INT8 when supported. Once converted, the engine is serialized and can be executed through the TensorRT runtime, which is tightly integrated with CUDA and NVIDIA’s kernel libraries. Compared to general-purpose libraries, TensorRT-LLM provides a lower-level execution model where computational and memory operations are highly optimized for modern GPUs (e.g., NVIDIA’s Hopper architecture).

Beyond these system-level design advantages, recent empirical evidence further motivates the use of TensorRT-LLM. In particular, the authors of \cite{niu2025eeorexaustive} present a comparative analysis of the frameworks mentioned above, and find that vLLM and TensorRT-LLM consistently achieve the most energy-efficient token generation among the compared frameworks, especially under high-concurrency and high-throughput workloads. 
Component-level measurements further show that both vLLM and TensorRT-LLM consume only about $5\%$ of the GPU energy compared to a general-purpose serving framework such as Transformers~\cite{huggingface-transformers}. This result highlights how their hardware-aware optimizations translate directly into power savings.
Collectively, these results show that TensorRT-LLM is not only engineered for performance but is also empirically validated as one of the most energy-efficient inference engines available. Based on these considerations, as well as to obtain a production-grade figure of merit, we choose TensorRT-LLM as the inference run-time in this paper. Future work will extend the analysis to different frameworks.

\subsubsection{\textbf{In-Flight Batching}}
A key run-time optimization employed by TensorRT-LLM to maximize throughput under dynamic workloads is its \emph{in-flight batching} strategy. Unlike static batching, where requests must be grouped prior to execution and processed synchronously, in-flight batching enables the run-time to continuously merge and schedule requests that are already undergoing inference. This mechanism is particularly well-suited for autoregressive LLM decoding, where sequences progress token by token and naturally diverge in length. At a high level, TensorRT-LLM decouples request admission from kernel execution. Incoming inference requests are first tokenized and then admitted into an active set of sequences maintained by the run-time scheduler. During each decoding iteration, TensorRT-LLM dynamically forms micro-batches composed of all active sequences that are ready to generate the next token. Newly arriving requests can be inserted into this active set without waiting for the completion of previous batches, while completed sequences are removed as soon as they reach their termination condition. As a result, GPU execution remains continuously occupied, reducing idle cycles that would otherwise arise from batch fragmentation. From a system perspective, in-flight batching trades strict per-request latency determinism for significantly improved throughput and hardware utilization, being especially effective in production serving scenarios, where request arrival times and sequence lengths are inherently unpredictable.

The behavior of in-flight batching is managed by user-configurable limits, namely \texttt{\seqsplit{max\_batch\_size}} and \texttt{\seqsplit{max\_num\_tokens}}. The \texttt{\seqsplit{max\_batch\_size}} parameter specifies the maximum number of active requests that can be scheduled concurrently in a decoding iteration, effectively bounding the degree of request-level parallelism. In contrast, \texttt{\seqsplit{max\_num\_tokens}} constrains the total number of tokens processed in a single micro-batch, aggregating across all active sequences. This token-level limit is particularly important for controlling GPU memory usage per iteration.

\section{Related Work}
\label{sec:related_works}
Several recent works have characterized and modeled the energy footprint of LLM inference. 
For instance, \cite{fernandez2025energyconsiderationslargelanguage} presents a systematic analysis of inference energy-efficiency optimizations across diverse NLP and generative AI workloads, considering input–output token distributions, batching strategies, GPU hardware, software frameworks, and decoding methods. Their work shows that real-world energy consumption is highly sensitive to workload geometry and software–hardware configurations. 
Authors of \cite{niu2025eeorexaustive} characterize the energy consumption of different inference engines (Transformers, vLLM, DeepSpeed, and TensorRT-LLM). They compute the Energy-per-Token metric $E_{\text{tok}}$ as:
\begin{equation}\label{eq:e_pt_n_out}
    E_{\text{tok}} = \frac{E_{\text{tot}}}{n_{\text{out}}}
\end{equation}
where $E_{\text{tot}}$ is the total energy consumption during inference and $n_{\text{out}}$ the total generated output tokens. In \cite{wilhelm2025beyondtestime}, they investigate the trade-offs between accuracy and energy consumption when applying test-time compute strategies. They propose an energy-aware routing mechanisms to guide sustainable model selection and inference, defining $E_{\text{tok}}$ to account for the number of input tokens $n_{\text{in}}$, by expressing it as:
\begin{equation}\label{eq:e_pt_n_out_n_in}
    E_{\text{tok}} = \frac{E_{\text{tot}}}{n_{\text{out}} + n_{\text{in}}}
\end{equation}
Finally, in \cite{wilkins2024offlineenergyoptimalllmserving} they characterize the energy consumption and run-time behavior of a set of LLMs using A100 GPUs, proposing workload-based energy models that,  a given model predicts consumption as a function of only the input and output tokens. To the best of our knowledge, this is the only work that proposes a model to describe the total energy consumption as a function of input and output token as follows:
\begin{equation}\label{eq:e_consumed_cambridge}
    E_{\text{tot}} = 
    \theta_{0}n_{\text{in}}
    + \theta_{1} n_{\text{out}}
    + \theta_{3} n_{\text{in}} n_{\text{out}}
\end{equation}
In contrast to prior studies, we introduce \textit{SweetSpot}, an analytical model of LLM inference energy consumption derived from both computational and memory-access complexity. This model enables us to characterize the non-linear interplay between input and output sequence lengths. To assess it, we conduct a systematic study of inference energy usage using TensorRT-LLM on NVIDIA H100 GPUs, across a diverse set of models and a wide range of sequence lengths. Our formulation makes explicit how quadratic input-processing costs and linear decoding costs combine to shape the efficiency patterns observed in practice, and we further extend the model to incorporate memory-access effects. Through extensive evaluation across multiple model families, we show that \textit{SweetSpot} closely matches measured energy trends and correctly identifies where efficiency peaks occur as a function of input and output lengths. Finally, we show that efficiency does not vary monotonically with sequence length; instead, it follows distinct regimes, with clear "sweet-spots" configurations and degradation zones depending on the balance between input size and output length.

\begin{table}
\scriptsize
\caption{Analytical Model Set}
\centering
\begin{tabularx}{\linewidth}{cX} 
\toprule

\textbf{Model} & \textbf{Formula} \\

\midrule

\textbf{\makecell[c]{Baseline 1 \\ \smaller \cite{niu2025eeorexaustive}}} & 
\( E_{\text{tok}} = 
\theta_{0}
\) \\

\midrule

\textbf{\makecell[c]{Baseline 2 \\ \smaller \cite{niu2025eeorexaustive}}} & 
\( E_{\text{tok}}( n_{\text{out}}) = 
\theta_{0}
+ \frac{\theta_{1}}{n_{\text{out}}}
\) \\

\midrule

\textbf{\makecell[c]{Baseline 3 \\ \smaller \cite{wilhelm2025beyondtestime}}} & 
\( E_{\text{tok}}(n_{\text{in}}, n_{\text{out}}) = 
\theta_{0}
+ \frac{\theta_{1}}{n_{\text{in}} + n_{\text{out}}}
\) \\

\midrule

\textbf{\makecell[c]{Baseline 4 \\ \smaller \cite{wilkins2024offlineenergyoptimalllmserving}}} & 
\( E_{\text{tok}}(n_{\text{in}}, n_{\text{out}}) = 
\theta_{0}
+ \frac{\theta_{1} n_{\text{in}}}{n_{\text{out}}}
+ \theta_{2} n_{\text{in}}
\) \\

\midrule

\textbf{\makecell[c]{SweetSpot (FLOPs-only) \\ \smaller (ours)}} &
\(
E_{\text{tok}}(n_{\text{in}}, n_{\text{out}}) =
  \theta_{0}
+ \frac{\theta_{1} n_{\text{in}}^{2}}{n_{\text{out}}}
+ \theta_{2} n_{\text{in}}
+ \frac{\theta_{3} n_{\text{in}}}{n_{\text{out}}}
+ \theta_{4} n_{\text{out}}
\) \\

\midrule

\textbf{\makecell[c]{SweetSpot\\ \smaller (ours)}} &
\(
E_{\text{tok}}(n_{\text{in}}, n_{\text{out}}) =
  \theta_{0}
+ \frac{\theta_{1} n_{\text{in}}^{2}}{n_{\text{out}}}
+ \theta_{2} n_{\text{in}}
+ \frac{\theta_{3} n_{\text{in}}}{n_{\text{out}}}
+ \theta_{4} n_{\text{out}}
+ \frac{\theta_{5}}{n_{\text{out}}}
\) \\

\bottomrule
\end{tabularx}
\label{tab:model_set}
\end{table}
\section{SweetSpot: Model Formulation and Energy Efficiency Prediction}
\label{sec:computation_model}
In this section, we present \textit{SweetSpot}, our analytical model for predicting LLM inference energy efficiency, explicitly analyzing the distinct computational characteristics of the prefill and decode phases. We first introduce a version of \textit{SweetSpot} based solely on the computational complexity of inference, and subsequently extend it to incorporate memory-access complexity. To assess the validity of our approach, we compare \textit{SweetSpot} against four baselines, summarized in \autoref{tab:model_set}.
From \cite{niu2025eeorexaustive} we derive Baseline 1, which simply describes the $E_{\text{tok}}$ as constant. Moreover, we formalize a parametrized model that relies solely on the number of generated tokens, $n_{\text{out}}$, to characterize $E_{\text{tok}}$, referred to as Baseline 2. 
The third baseline, Baseline 3, expands on this by also including the number of input tokens, $n_{\text{in}}$, as introduced in \autoref{eq:e_pt_n_out_n_in}. This formulation allows us to establish a relationship between $E_{\text{tok}}$ and the total sequence length. 
Finally, Baseline 4 is derived from \autoref{eq:e_consumed_cambridge}, normalizing the equation by $n_{\text{out}}$.

\subsection{SweetSpot Model (FLOPs-only)}
To evaluate the energy efficiency of our LLMs, we derived an initial analytical model grounded exclusively in the computational complexity of inference. The model accounts for both the \textit{prefill} phase, where the entire input sequence is processed in parallel, and the \textit{decode} phase, where tokens are generated autoregressively using the cached KV states. In the following, we present the derivation of the formulas for each phase and then combine them into a unified expression.
During the prefill stage, the model processes an input sequence of length $n_{\text{in}}$ in parallel. For a single Transformer layer with hidden size $d$, the computational cost can be decomposed into two main contributions: the Self-Attention Mechanism and the Feed-Forward Network (FFN). 
We neglect the cost of scaling, softmax, layer normalization, dropout, and bias additions, which are negligible w.r.t. the cost of attention and FFN.

Moreover, we adopt the standard Floating-point Operations (FLOPs), denoted by $F$, accounting convention where one multiply–accumulate operation is counted as two FLOPs (one multiplication and one addition).

\subsubsection{\textbf{Prefill Stage FLOPs}}
\paragraph{\textbf{Attention}}
The attention mechanism involves the following matrix multiplications:

Input projections to Query, Key, Value ($Q$, $K$, $V$): the input $X \in \mathbb{R}^{n_{\text{in}} \times d}$ is multiplied by three weight matrices $W_Q, W_K, W_V \in \mathbb{R}^{d \times d}$:
\[
    F_{\text{proj}} = 3 \cdot (2 n_{\text{in}} d^2) = 6 n_{\text{in}} d^2
\]
This corresponds to three dense matrix multiplications of size $n_{\text{in}} \times d$ by $d \times d$.
    
Attention score computation $QK^\top$: multiplication of $Q \in \mathbb{R}^{n_{\text{in}} \times d}$ with $K^\top \in \mathbb{R}^{d \times n_{\text{in}}}$:
\[
    F_{QK^T} = 2 n_{\text{in}}^2 d
\]  
Weighted sum with $V$: multiplication of the $n_{\text{in}} \times n_{\text{in}}$ attention matrix with $V \in \mathbb{R}^{n_{\text{in}} \times d}$:
\[
    F_V = 2 n_{\text{in}}^2 d
\]
Final output projection: multiplication with $W_O \in \mathbb{R}^{d \times d}$:
\[
    F_{\text{out}} = 2 n_{\text{in}} d^2
\]
Summing these terms yields the attention cost per layer:
\[
    F_{\text{att}} = 8 n_{\text{in}} d^2 + 4 n_{\text{in}}^2 d
\]
\paragraph{\textbf{FFN}}
The feed-forward network expands the hidden dimension by a factor of 4, with two dense matrix multiplications per token:
\[
F_{\text{FFN}} = 8 n_{\text{in}} d^2 + 8 n_{\text{in}} d^2 = 16 n_{\text{in}} d^2
\]
corresponding to the attention output multiplication with $W_1$ and the subsequent $W_2$ projection back to dimension $d$.
\paragraph{\textbf{Total Prefill}}
For one layer:
\[
\begin{aligned}
    F_{\text{prefil, layer}}=F_{\text{att}}+ F_{\text{FFN}}
    = 24 n_{\text{in}} d^2 + 4 n_{\text{in}}^2 d
\end{aligned}
\]
For an $L$-layer model:
\[
    F_{\text{prefill}} = L \cdot \big(24 n_{\text{in}} d^2 + 4 n_{\text{in}}^2 d \big)
\]
\subsubsection{\textbf{Decode Stage FLOPs}}
At decoding step $t$, the model has access to $n_{\text{in}} + t - 1$ cached tokens. For each new token, the per-layer costs are:

\paragraph{\textbf{Attention}}
Projections for $Q, K, V$, and output: multiplications with $d \times d$ weight matrices:
\[
    F_{\text{proj+out}} = 8 d^2
\]  
Attention with cached keys and values: multiplication of $Q \in \mathbb{R}^{1 \times d}$ with cached $K^\top \in \mathbb{R}^{d \times (n_{\text{in}} + t - 1)}$ and multiplication of resulting attention scores with $V \in \mathbb{R}^{(n_{\text{in}} + t - 1) \times d}$:
\[
    F_{QK^T+V}(t) = 4 (n_{\text{in}} + t - 1) d
\]
\paragraph{\textbf{FFN}}
Per token, the FFN again involves two $d \times d$ multiplications:
\[
    F_{\text{FFN}} = 16 d^2
\]
\paragraph{\textbf{Total per Token}}
The per-token cost for one layer is therefore:
\[
\begin{aligned}
    F_{\text{decode, layer}}(t) &=F_{\text{proj+out}} + F_{QK^T+V} + F_{\text{FFN}}
    \\&= 24 d^2 + 4 (n_{\text{in}} + t - 1) d
\end{aligned}
\]
\paragraph{\textbf{Total Decode}}
Summing over $n_{\text{out}}$ generated tokens and multiplying by $L$ layers:
\[
\begin{aligned}
    F_{\text{decode}} &= L \sum_{t=1}^{n_{\text{out}}} \big(24 d^2 + 4 (n_{\text{in}} + t - 1) d \big) \\
    &= L \big(24 n_{\text{out}} d^2 + 4 d (n_{\text{out}} n_{\text{in}} + \tfrac{n_{\text{out}}(n_{\text{out}}-1)}{2}) \big)
\end{aligned}
\]
\subsubsection{\textbf{Total Inference FLOPs}}
By summing the Prefill and Decode FLOPs, we obtain:
{
\smaller
\[
\begin{aligned}
    F_{\text{total}} &= F_{\text{prefill}} + F_{\text{decode}}
    =L \Big[ 24 d^2 (n_{\text{in}} + n_{\text{out}}) 
    + 4 d \big( n_{\text{in}}^2 + n_{\text{out}} n_{\text{in}} + \tfrac{n_{\text{out}}(n_{\text{out}}-1)}{2} \big) \Big]
\end{aligned}
\]
}
We factor out the common multiplier \(2Ld\) and manipulate the formula to spot individual contributors. This compact form is convenient for later normalization to build the final model.
{\small
\[
F_{\text{total}}
= 2\,L\,d \;\Big( 12\,d\,n_{\text{in}}
+ 12\,d\,n_{\text{out}}
+ 2\,n_{\text{in}}^{2}
+ 2\,n_{\text{out}}\,n_{\text{in}}
+ n_{\text{out}}^{2}
- n_{\text{out}} \Big)
\]
}
We note that the external factors \textit{L} and \textit{d} depends only on model size (depth and width) and are constant for a given model. Since our goal is to capture how efficiency varies as a function of input length \(n_{\text{in}}\) and output length \(n_{\text{out}}\), we remove these constant multipliers and express the energy as a linear combination of polynomial terms in terms of \(n_{\text{in}}\) and \(n_{\text{out}}\). Aggregating similar terms, we obtain:
{\small
\[
E_{\text{tot}}(n_{\text{in}},n_{\text{out}})
= \theta_{0}\, n_{\text{out}}
+ \theta_{1}\, n_{\text{in}}^{2}
+ \theta_{2}\, n_{\text{out}}\,n_{\text{in}}
+ \theta_{3}\, n_{\text{in}}
+ \theta_{4}\, n_{\text{out}}^{2}
\]
}
This formulation highlights the individual contributions of input length, output length, and their interaction. The learned parameters \(\theta_i\) capture the proportionality between the theoretical FLOP terms and the actual measured energy in Joules.

To analyze the computational cost on a \textit{per-output-token} basis, we divide the total energy by the number of generated tokens \(n_{\text{out}}\). This provides the average energy required to produce a single token, which is useful for comparing efficiency across different input–output configurations.
{\small
\begin{equation}\label{eq:flop_model}
E_{\text{tok}}(n_{\text{in}}, n_{\text{out}})
= \theta_{0}
+ \frac{\theta_{1}\, n_{\text{in}}^{2}}{n_{\text{out}}} 
+ \theta_{2}\, n_{\text{in}}
+ \frac{\theta_{3}\, n_{\text{in}}}{n_{\text{out}}} 
+ \theta_{4}\, n_{\text{out}}
\end{equation}
}

This normalized form highlights how the cost of processing the input sequence \(n_{\text{in}}\) is amortized across the output tokens, while the terms proportional to \(n_{\text{out}}\) reflect the per-token cost of decoding. 
\subsection{SweetSpot Model}
\label{sec:computation_and_memory_model}
While FLOPs capture the arithmetic workload of inference, modern accelerators increasingly operate in a regime where energy consumption is dominated not by computation but by memory traffic, in particular, the movement of activations, weights, and KV-cache tensors across the memory hierarchy \cite{llminfrence_unveiled}. To accurately model energy, we therefore incorporate the memory-access cost alongside the computational cost derived in the previous section. In the following subsections, we present an analytical decomposition of the memory contribution during both the prefill and decode phases, denoted by $M$. As before, we analyze a single Transformer layer with hidden dimension $d$ and input length $n_{\text{in}}$, and $n_q$ number of heads. We again omit the negligible contributions of bias reads, layer-normalization reads/writes, and softmax.

\subsubsection{\textbf{Prefill Stage Memory Operations}}
During prefill, the model processes the entire sequence in parallel, producing input projections and attention tensors. Memory traffic arises from reading inputs, writing intermediate projected representations and attention matrices. For one layer, the total memory movement is decomposed as follows.
\paragraph{\textbf{Attention}}
The attention mechanism involves the following memory transfers:\\
Input projections to $Q$, $K$, $V$: reading $X \in \mathbb{R}^{n_{\text{in}} \times d}$ and writing projected tensors:
\[
    M_{\text{proj}} = 2 n_{\text{in}} d + d^{2}
\]
Attention score computation $QK^\top$: reading $Q \in \mathbb{R}^{n_{\text{in}} \times d}$ and $K^\top \in \mathbb{R}^{d \times n_{\text{in}}}$ and writing the $n_{\text{in}} \times n_{\text{in}}$ attention matrix:
\[
    M_{QK^T} = 2 n_{\text{in}} d + n_{\text{in}}^{2} n
\]
Weighted sum with $V$: reading the attention matrix and $V \in \mathbb{R}^{n_{\text{in}} \times d}$ ($n_q$ = number of heads):
\[
    M_{V} = 2 n_{\text{in}} d + n_{\text{in}}^{2} n_q
\]
Final output projection: reading the attention output ($n_{\text{in}} \times d$) and writing the output representation:
\[
    M_{\text{out}} = 2 n_{\text{in}} d + d^{2}
\]
Summing these terms gives the total attention memory per layer during prefill:
\[
M_{\text{att}} = 8 n_{\text{in}} d + 2 d^2 + 2 n_{\text{in}}^2 n_q
\]
\paragraph{\textbf{FFN}}
The feed-forward network reads the input activations, accesses its two weight matrices, and writes the intermediate expanded representation:
\[
M_{\text{FFN}} = 2 n_{\text{in}} d + 8 d^2
\]
\paragraph{\textbf{Total Prefill}}
For one layer:
\[
M_{\text{prefil, layer}} = M_{\text{att}} + M_{\text{FFN}} = 10 n_{\text{in}} d + 10 d^2 + 2 n_{\text{in}}^2 n_q
\]
For an $L$-layer model:
\[
M_{\text{prefill}} = L \left( 10 n_{\text{in}} d + 10 d^2 + 2 n_{\text{in}}^2 n_q\right)
\]
\subsubsection{\textbf{Decode Stage Memory Operations}}
Similarly to the FLOPs Decoding stage, at each step $t$, we describe the per-layer memory operations as follows.\\
\paragraph{\textbf{Attention}}
Projections for $Q, K, V$, and output: multiplications with $d \times d$ weight matrices:
\[
    M_{\text{proj+out}} = 2(2d + d^2)
\]   
Attention with cached keys and values: multiplication of $Q \in \mathbb{R}^{1 \times d}$ with cached $K^\top \in \mathbb{R}^{d \times (n_{\text{in}} + t - 1)}$ and multiplication of resulting attention scores with $V \in \mathbb{R}^{(n_{\text{in}} + t - 1) \times d}$:
\[
    M_{QK^T+V}(t) = 2 ((n_q + d)(n_{\text{in}} + t - 1) + d)
\]
\paragraph{\textbf{FFN}}
Per token, the FFN again involves two $d \times d$ multiplications:
\[
    M_{\text{FFN}} = 2d + 8d^2
\]
\paragraph{\textbf{Total per Token}}
The per-token cost for one layer is therefore:
\[
\begin{aligned}
    M_{\text{decode, layer}}(t) &= M_{\text{proj+out}} + M_{QK^T+V} + M_{\text{FFN}} \\
    &= 7d + 10d^2 + (n_q + d)(n_{\text{in}} + t - 1)
\end{aligned}
\]
\paragraph{\textbf{Total Decode}}
Summing over $n_{\text{out}}$ generated tokens and multiplying by $L$ layers:
{\small
\[
\begin{aligned}
    M_{\text{decode}} &= L \sum_{t=1}^{n_{\text{out}}} \big(7d + 10d^2 + (n_q + d)(n_{\text{in}} + t - 1)\big) \\
    &= L \big(7dn_{\text{out}} + 10 d^2n_{\text{out}} + (n_q + d)(n_{\text{in}}n_{\text{out}} + \tfrac{n_{\text{out}}(n_{\text{out}}-1)}{2}) \big)
\end{aligned}
\]
}
\subsubsection{\textbf{Total Memory Operations}}
By summing the Prefill and Decode Memory Operations, we obtain:
{\small
\[
\begin{aligned}
    M_{\text{total}} &=
    M_{\text{prefill}} + M_{\text{decode}} \\
    &= L \Big[10 n_{\text{in}} d + 10 d^2 + 2 n_{\text{in}}^2 n_q \\
    &\quad + n_{\text{out}}\big(7d + 10 d^2 + (n_q + d)(n_{\text{in}} + \tfrac{(n_{\text{out}}-1)}{2}\big) \Big]
\end{aligned}
\]
}
As for the FLOPs-only model, we express the energy consumption as a linear combination of polynomial terms in $n_{\text{in}}$ and $n_{\text{out}}$:
{\small
\[
E_{\text{tot}}(n_{\text{in}},n_{\text{out}})
= \theta_{0}\, n_{\text{out}}
+ \theta_{1}\, n_{\text{in}}^{2}
+ \theta_{2}\, n_{\text{out}}\,n_{\text{in}}
+ \theta_{3}\, n_{\text{in}}
+ \theta_{4}\, n_{\text{out}}^{2}
+ \theta_{5}
\]
}
Relative to the $E_{\text{tot}}$ of the FLOPs model, this formulation introduces an additional constant term, $\theta_{6}$. This term accounts for the contribution arising from the $10 d^{2}$ component in the $M_{\text{prefill}}$ expression, which is independent of both $n_{\text{in}}$ and $n_{\text{out}}$.
Analyzing the energy consumption \textit{per-output-token}, we obtain:
{\small
\begin{equation}\label{eq:flop_mem_model}
E_{\text{tok}}(n_{\text{in}}, n_{\text{out}})
= \theta_{0}
+ \frac{\theta_{1}\, n_{\text{in}}^{2}}{n_{\text{out}}} 
+ \theta_{2}\, n_{\text{in}}
+ \frac{\theta_{3}\, n_{\text{in}}}{n_{\text{out}}} 
+ \theta_{4}\, n_{\text{out}}  
+ \frac{\theta_{5}}{n_{\text{out}}}
\end{equation}
}

\subsection{Peak Energy Efficiency Prediction}
Using the formulation of \textit{SweetSpot} introduced in \autoref{eq:flop_mem_model}, we identify peak energy-efficiency configurations by considering the per-token energy as a function of the number of generated tokens~$n_{\text{out}}$.  
To this end, we compute the derivative of $E_{\text{tok}}$ with respect to $n_{\text{out}}$ and set it equal to zero:
\[
\frac{\partial E_{\text{tok}}}{\partial n_{\text{out}}} 
= - \frac{\theta_1 n_{\text{in}}^2 + \theta_3 n_{\text{in}} + \theta_5}{n_{\text{out}}^2} + \theta_4 = 0
\]

Solving for \(n_{\text{out}}\), we obtain:
\[
\frac{\theta_1 n_{\text{in}}^2 + \theta_3 n_{\text{in}} + \theta_5}{n_{\text{out}}^2} = \theta_4
\quad \Rightarrow \quad
n_{\text{out}}^2 = \frac{\theta_1 n_{\text{in}}^2 + \theta_3 n_{\text{in}} + \theta_5}{\theta_4}
\quad \Rightarrow \quad
\]
\begin{equation}\label{eq:sweet-spots}
\boxed{n_{\text{out}}^\ast = \sqrt{\frac{\theta_1 n_{\text{in}}^2 + \theta_3 n_{\text{in}} + \theta_5}{\theta_4}}}
\end{equation}
This result shows that the per-token energy is minimized when the number of generated tokens is equal to the square root of a combination of the input sequence length and its quadratic contribution, normalized by the linear per-token decoding cost \(\theta_4\).
\section{Experimental Study}
\label{sec:experimental_study}
This section details the experimental study conducted to validate \textit{SweetSpot} and to evaluate the energy consumption of LLMs inference on modern hardware. We describe the experimental design, including the monitoring framework, LLM selection, metrics, datasets, engine configurations, benchmarking procedure, and system setup. We then report and analyze the experimental results, evaluating the energy efficiency of the LLM test set and the accuracy of our proposed analytical model \textit{SweetSpot}.
\subsection{Experimental Design}
This subsection formalizes the experimental protocol, describing the measurement pipeline and design choices adopted to ensure controlled and reproducible evaluation across all experiments.
\subsubsection{\textbf{Monitoring Framework}}
\label{subsec:monitoring_framework}
To collect and analyze GPU metrics during LLM inference, we employ an updated version of ExaMon \cite{bartolini2019examon}, based on IoTdb Database \cite{iotdb}.
In particular, hardware metrics such as \texttt{\seqsplit{nvmlDeviceGetPowerUsage}} and \texttt{NVML\_CLOCK\_GRAPHICS} are sampled using the NVIDIA Management Library Python API (pynvml \cite{pynvml}). These metrics are sampled at \textit{500 ms} intervals, published through the ExaMon plugin, and stored in the ExaMon database. 

\subsubsection{\textbf{LLM Selection}}
\begin{table}
\small
\centering
\caption{LLM test set grouped into size-based families (1B–9B parameters), with architectural details for each model.}
\begin{tabularx}{\columnwidth}{Xcccc}
\toprule
\textbf{Name} & \textbf{Size (B)} & \textbf{Heads-Attention} & \textbf{Hidd. Size} & \textbf{Layers} \\
\midrule
\multicolumn{5}{l}{\textbf{XS models (From 1B to 1.5B)}} \\
Llama 3.2   & 1   & 32 - GQA (4x) & 2048 & 16 \\
OPT         & 1.3 & 32 - MHA      & 2048 & 24 \\
Qwen 2      & 1.5 & 12 - GQA (6x) & 1536 & 28 \\ [4pt]

\multicolumn{5}{l}{\textbf{S models (From 2B to 3B)}} \\
Gemma 2     & 2   & 8 - GQA (2x)  & 2304 & 26 \\
OPT         & 2.7 & 32 - MHA      & 2560 & 32 \\
Llama 3.2   & 3   & 24 - GQA (4x) & 3072 & 28 \\
Granite     & 3   & 32 - MHA      & 2048 & 32 \\ [4pt]

\multicolumn{5}{l}{\textbf{M models (From 6.7B to 9B)}} \\
OPT         & 6.7 & 32 - MHA      & 4096 & 32 \\
Qwen 2      & 7   & 28 - GQA (7x) & 3584 & 28 \\
Falcon-RW   & 7 & 64 - MHA        & 4096 & 36 \\
Granite     & 8   & 32 - GQA (4x) & 4096 & 36 \\
Llama 3.1   & 8   & 32 - GQA (4x) & 4096 & 32 \\
Gemma 2     & 9   & 16 - GQA (2x) & 3584 & 42 \\ [4pt]

\bottomrule
\label{tab:llm_set}
\end{tabularx}
\end{table}
We consider a diverse set of LLMs, ranging from 1B to 9B parameters, as summarized in \autoref{tab:llm_set}. Beyond model size, the selection includes architectures with different design choices and attention mechanisms, including Multi-Head Attention (MHA) and Grouped-Query Attention (GQA). For GQA, we also account for different KV grouping ratios e.g. GQA (4×) denotes four query heads sharing a single KV head. To facilitate comparison, the models are grouped into three size-based families, ranging from XS models (1B–1.5B parameters) to M models (6.7B–9B parameters).

\subsubsection{\textbf{Metrics and Measurement}}
We quantify the energy consumption $E_{\text{tot}}$ as the discrete integration of the mean instantaneous GPU power, measured as described in Section \ref{subsec:monitoring_framework} over the measurement interval $\Delta t$, according to the following expression: 
\[
E_{\text{tot}} = \sum_i{\left(\frac{P_i + P_{i+1}}{2}\right) \Delta t}
\]
where $P_i$ denotes the power usage recorded at each time step $i$.
We empirically obtain the Energy-per-Token ($E_{\text{tok}}$) by normalizing the energy consumption by the total amount of output tokens $n_{\text{out}}$:
\[
E_{\text{tok}} = \frac{E_{\text{tot}}}{n_{\text{out}}} \quad \left[\frac{\mathrm{\text{Joules}}}{\mathrm{\text{Token}}}\right]
\]
For improved interpretability, we additionally define the LLM energy efficiency as the number of output tokens generated per unit of energy. Formally, this metric corresponds to the reciprocal of $E_{\text{tok}}$:
\[
E_{\text{eff}} = \frac{n_{\text{out}}}{E_{\text{tot}}} \quad \left[\frac{\mathrm{\text{Tokens}}}{\mathrm{\text{Joule}}}\right]
\]
This measure captures the productivity of the model with respect to its energy budget, facilitating comparisons between models.

\subsubsection{\textbf{Dataset Generation}}
To benchmark TensorRT-LLM under controlled and reproducible conditions, we generated synthetic request datasets using the official utility script provided with TensorRT-LLM. For each dataset, token sequences were sampled using the Llama-3.1-8B-Instruct tokenizer, with input and output lengths drawn from fixed distributions defined by a specified mean and zero variance. This allowed us to precisely control the number of input and output tokens per request. We generated the dataset.\footnote{The dataset is publicly available on Zenodo: \href{https://doi.org/10.5281/zenodo.18714476}{doi.org/10.5281/zenodo.18714476}} for all combinations of input and output lengths ranging from 64 to 4096 tokens (in powers of two), and for different workload sizes with the number of requests set to 10, 100, and 1000. 
\subsubsection{\textbf{Engines Configuration}}
All TensorRT-LLM engines used in our benchmarks were built using the \texttt{trtllm-bench build}, which converts model checkpoints into TensorRT-LLM engines. Engines were constructed with tensor parallelism and pipeline parallelism both set to one, ensuring single-GPU execution and isolating inference performance from inter-device communication effects. The maximum supported sequence length was set to 8192 tokens, allowing all benchmarked input–output combinations to execute within a single engine configuration without triggering engine rebuilds. Additionally, we configured the \texttt{\seqsplit{max\_batch\_size}} to 1024 and the \texttt{\seqsplit{max\_num\_tokens}} to 8192. This approach ensures that performance differences observed across experiments are attributable to workload characteristics rather than changes in engine structure or parallelization strategy.
\subsubsection{\textbf{Benchmark Execution}}
Benchmark runs were executed with \texttt{trtllm-bench} in \textit{throughput} mode, a dedicated TensorRT-LLM utility for evaluating LLM performance on a given dataset, designed to maximize sustained token generation rate under a fixed workload. Each run pairs a prebuilt engine with a corresponding synthetic dataset, ensuring that model configuration and workload characteristics remain constant throughout the measurement. We first assessed different number of requests (10, 100, and 1000) using Falcon-RW 7B. Subsequently, we evaluated the LLMs listed in \autoref{tab:llm_set} by fixing the number of requests to 1000, while varying the input and output sequence lengths from 64 to 4096 tokens. By reusing the same engine across datasets and varying only the dataset parameters, the reported results isolate the impact of input/output length and request count on inference throughput.
\subsubsection{\textbf{System Setup}}
All experiments were conducted on a single-node server equipped with an \textit{Intel\textsuperscript{\textregistered} Xeon\textsuperscript{\textregistered} Platinum 8480+} processor, belonging to the Sapphire Rapids family. In addition, the system hosts four NVIDIA H100 Tensor Core GPUs, connected through NVLink. 
The server is equipped with 32x64 GB of DDR5 RAM operating at a frequency of 4800~MHz. 
The system runs on a \textit{Dell 00G41X Version A02} motherboard. The operating system is Alma Linux 9.5 with Linux kernel 5.14.
The cooling infrastructure relies on a rack-mounted air cooling system with redundant fans, which operated at stable speeds of  5160~RPM during the experiments.
\begin{figure*}
    \centering

    \subfloat[$N_{\text{req}} = 10$.\label{fig:heatmap_requests10}]{
        \includegraphics[width=0.3\textwidth]{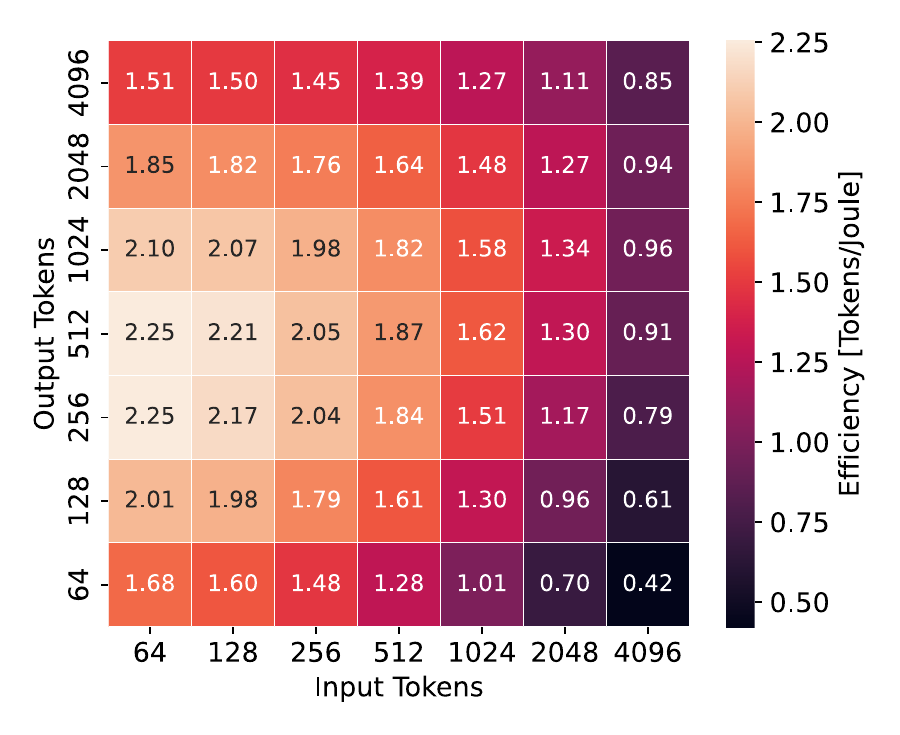}
    }\hfill
    \subfloat[$N_{\text{req}} = 100$.\label{fig:heatmap_requests100}]{
        \includegraphics[width=0.3\textwidth]{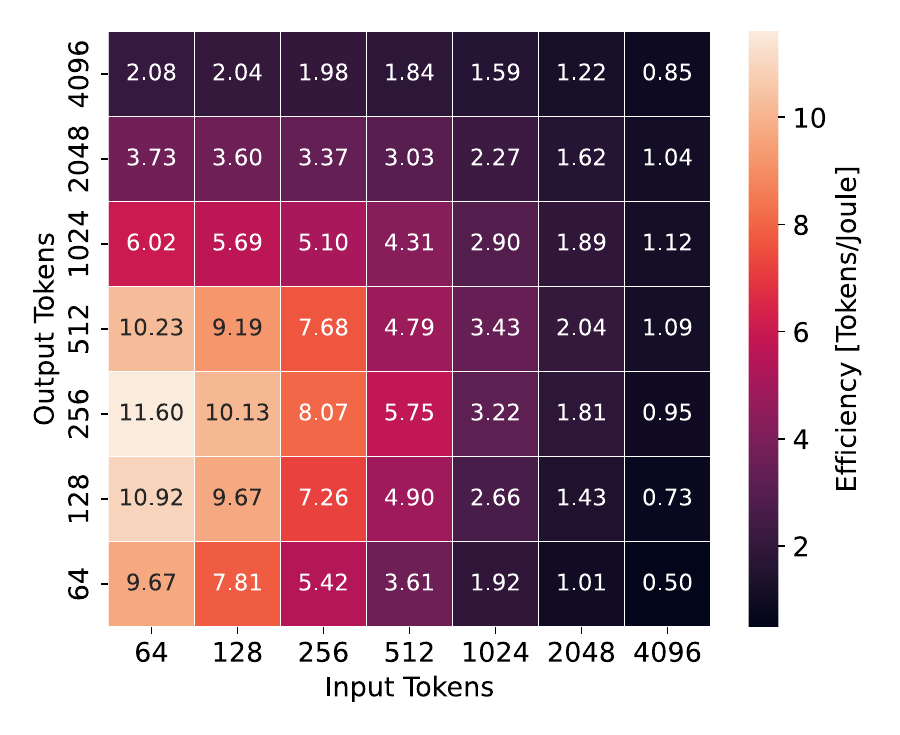}
    }\hfill
    \subfloat[$N_{\text{req}} = 1000$.\label{fig:heatmap_requests1000}]{
        \includegraphics[width=0.3\textwidth]{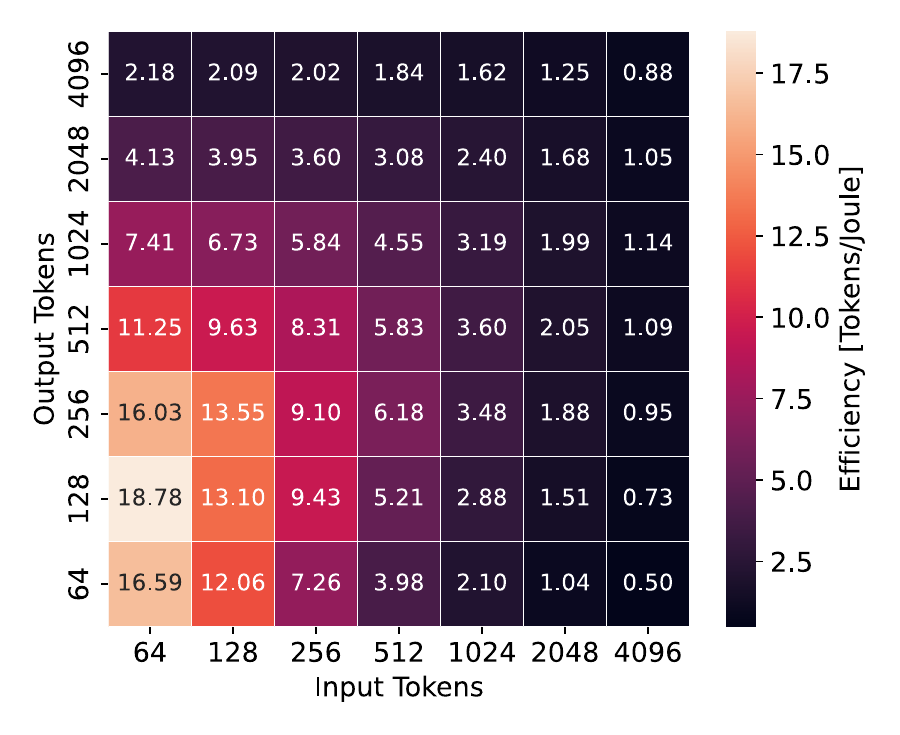}
    }

    \caption{Energy efficiency $\frac{\text{Tokens}}{\text{Joule}}$ of Falcon-RW 7B with different numbers of requests (10, 100, 1000).}
    \Description{Three heatmaps showing the energy efficiency, measured as tokens per joule, of Falcon-RW 7B for 10, 100, and 1000 requests.}
    \label{fig:heatmaps_requests}
\end{figure*}
\begin{figure}
    \centering
\includegraphics[width=\columnwidth]{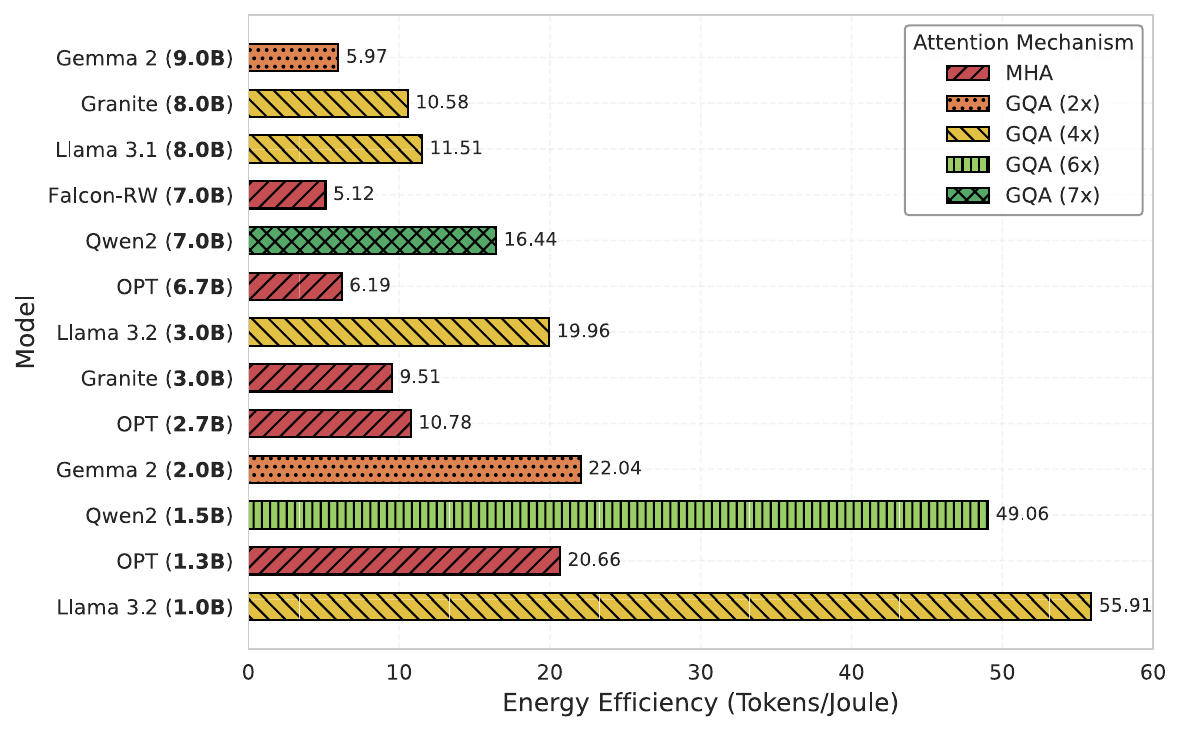}
    \caption{LLMs average Energy Efficiency across the entire dataset.}
    \Description{LLMs average Energy Efficiency across the entire dataset.}
    \label{fig:average_efficiency}  
\end{figure}
\subsection{Experimental Results}
In this subsection we discuss the results of our experimental study, evaluating the energy efficiency of the tested LLMs and the accuracy of our analytical model \textit{SweetSpot}, using it to predict the peak energy efficiency spots.
\label{subsec:experimental_results}
\begin{figure*}[t]
    \centering
    \subfloat[XS size models.\label{fig:efficiency_xs}]{
        \includegraphics[width=0.3\textwidth]{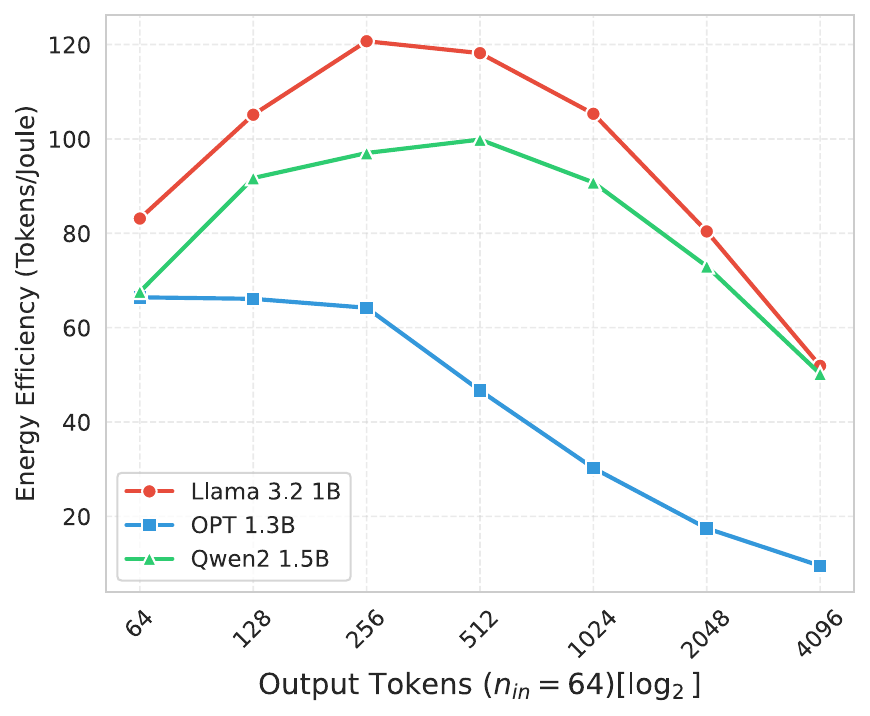}
    }\hfill
    \subfloat[S size models.\label{fig:efficiency_s}]{
        \includegraphics[width=0.3\textwidth]{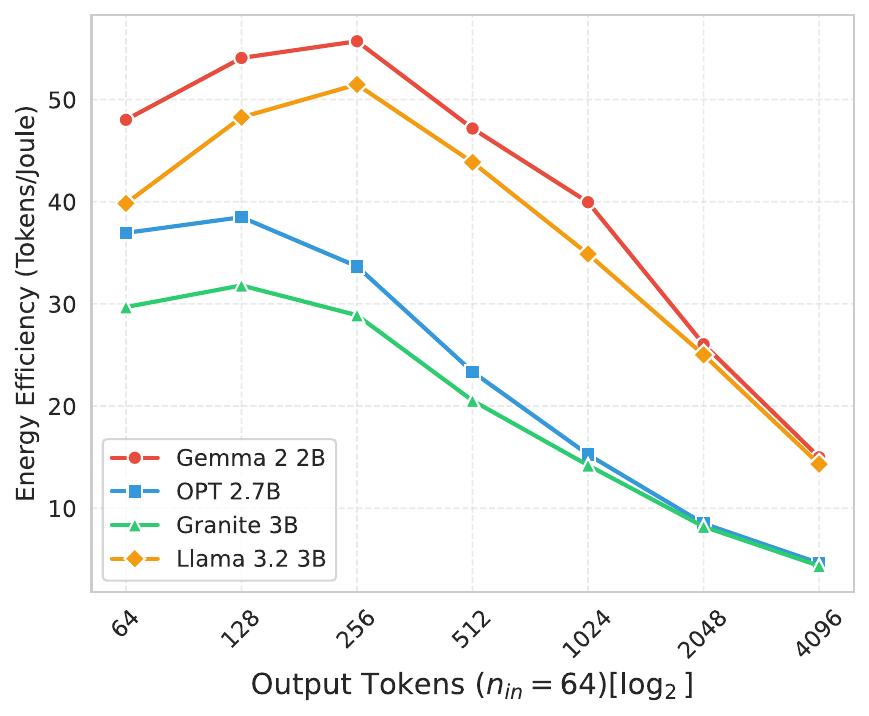}
    }\hfill
    \subfloat[M size models.\label{fig:efficiency_m}]{
        \includegraphics[width=0.3\textwidth]{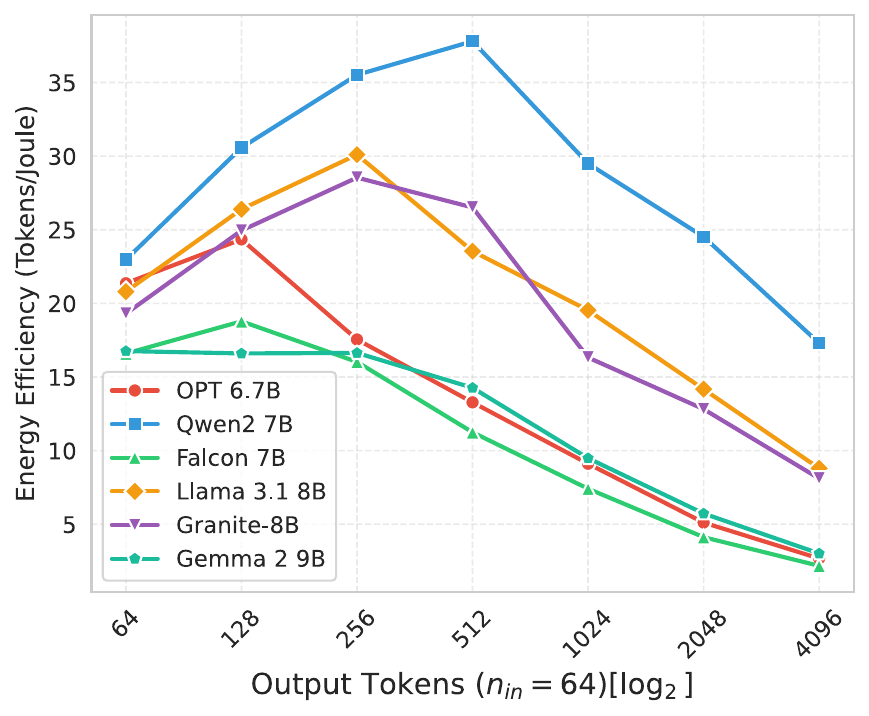}
    }
    \caption{Energy efficiency of tested LLMs considering $n_{\text{in}}=64$ and varying output tokens, per size family (XS, S, M).}
    \Description{Three line plots showing energy efficiency of XS, S, and M size LLMs with fixed $n_{\text{in}}=64$ and varying output tokens.}
    \label{fig:lineplot}
\end{figure*}
\subsubsection{\textbf{Number of Requests Selection}}
To assess the impact of the request count $N_{\text{req}}$ (i.e. the number of queries used for evaluation) on energy efficiency, we considered multiple workload sizes. We excluded the single-request setting because our monitoring framework introduces a non-negligible timing error, preventing reliable measurements in that regime where execution times are shorter.
\autoref{fig:heatmaps_requests} shows the energy efficiency for the Falcon-RW 7B model (the least efficient LLM of our set) across different workload size (10,100 and 1000). Each plot reports the average energy efficiency for different LLM inference run, each with a different $n_{\text{out}}$ (x-axis) and $n_{\text{in}}$ (y-axis) configurations. As shown in the figure, the workload size strongly affects both the absolute efficiency values and their overall trends. In particular, the efficiency peak shifts toward larger output lengths when using fewer requests, a behavior driven by the reduced number of tokens processed per inference step. Furthermore, the spread between the highest and the lowest efficiency values narrows for smaller workloads. Specifically, with 10 requests, $E_{\text{eff}}$ ranges from 0.42 to 2.25, which consists in an increase of about 6$\times$. In contrast, with 1000 requests, the values range from 0.50 to 18.78, corresponding to a significantly wider gap of approximately $37.6\times$.

Finally, smaller workloads amplify the relative impact of measurement overhead and necessitate shorter sampling intervals, which can further amplify measurement noise and result in less reliable estimates. For these reasons, in the subsequent experiments we focus on the saturation scenario and fix $N_{\text{req}}$ to 1000. A more detailed analysis of how workload size interacts with batch size is provided in Section \ref{subsec:batch-size_req}.
\subsubsection{\textbf{Overall LLMs Energy Efficiency}}
In \autoref{fig:average_efficiency}, we report the average energy efficiency measured across the entire dataset for each LLM in our test (see \autoref{tab:llm_set}), highlighting the attention mechanism adopted by each model. Among the evaluated models, Llama 3.2 1B achieves the highest overall energy efficiency, with an average score of 55.91. In contrast, Falcon-RW 7B exhibits the lowest efficiency, with an average value of 5.12.

Model size emerges as a primary factor influencing energy efficiency. In general, larger models tend to be less energy efficient than smaller ones. However, parameter count alone does not fully explain the observed differences. The attention mechanism also plays a significant role. In particular, models employing MHA are consistently less energy efficient than comparably sized models adopting GQA. Unlike MHA, where each attention head maintains independent Key–Value (KV) projections, GQA allows multiple query heads to share the same KV projections, which reduces the size of the KV cache and the number of memory accesses required at each decoding step, leading to lower bandwidth pressure and improved energy efficiency.
For example, OPT 1.3B (employing MHA) is approximately 42\% less efficient than Qwen2 1.5B (GQA 6x), achieving average efficiencies of 20.66 and 49.06, respectively, despite having fewer parameters. A similar pattern is observed when comparing OPT 6.7B and Qwen2 7B.
Moreover, higher GQA grouping ratios appear to correlate with improved efficiency. For instance, Qwen 7B (GQA 7x) is only about 18\% less efficient than Llama 3.2 3B (GQA 4x), despite having more than twice the number of parameters. This suggests that architectural choices in the attention mechanism can partially mitigate the energy cost typically associated with increased model size.

While the attention mechanism is a critical determinant of energy efficiency, it is not the only architectural factor involved. The number of layers, hidden dimensionality, intermediate projection sizes, and other architectural design choices also substantially impact overall efficiency. A systematic analysis of how these architectural components interact with energy consumption is left for future work.
\subsubsection{\textbf{Observation of Peak Energy Efficiency Spots}}
In \autoref{fig:lineplot}, we show the energy efficiency of each LLM in our test set, reported in \autoref{tab:llm_set}, considering a fixed input length of $n_{\text{in}} = 64$, the input length which shows the maximum efficiency, and varying the output length. 
The sub-figures correspond to different LLM size families (XS, S, M), and illustrate how energy efficiency (y-axis) varies with the output length $n_{\text{out}}$ (x-axis). 
We see that different LLMs exhibit a substantial spread in $E_{\text{eff}}$. For instance, LLama 3.2 1B is the most efficient LLM, reaching 120.71 at $n_{\text{out}} = 256$ and 51.90 at $n_{\text{out}} = 4096$, whereas Falcon-RW 7B is the least efficient, with energy efficiency values of 18.78 and 2.18 at the same output lengths.
While the absolute efficiency values differ across models due to architectural differences, the overall behavior is consistent: each model exhibits a non-linear curve with a distinct peak in efficiency.

\begin{figure}
    \centering
\includegraphics[width=0.7\columnwidth]{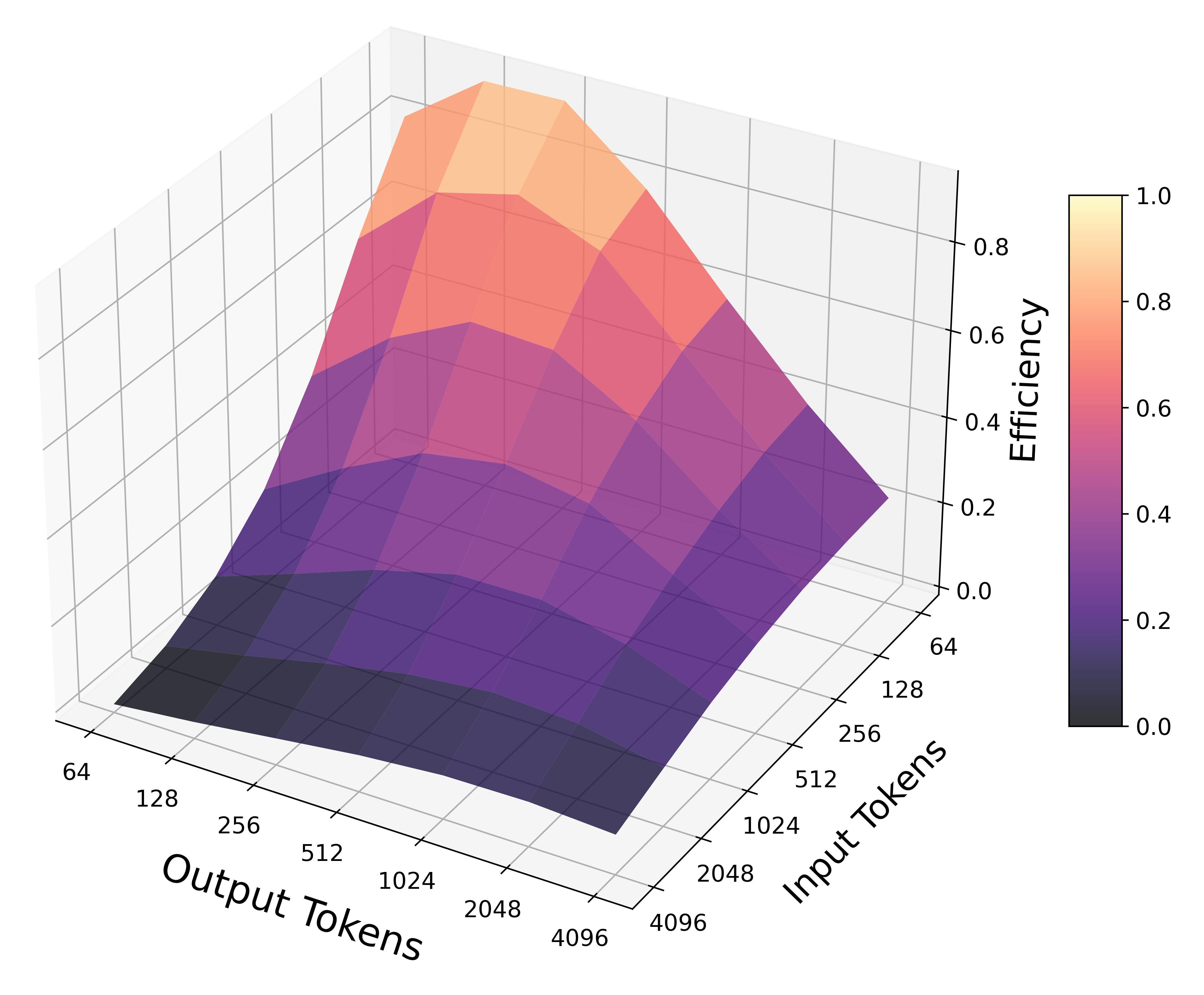}
    \caption{Aggregated energy efficiency heatmap.}
    \Description{Heatmaps showing the model Energy Efficiency aggregated average.}
    \label{fig:heatmap_average}  
\end{figure}

To highlight this common trend, we construct an aggregated heatmap in  \autoref{fig:heatmap_average} by applying min–max normalization to the efficiency values of each LLM and then averaging across the full set.  The resulting 3D visualization reports the energy efficiency (z-axis), bounded between 0 and 1, and illustrates how it varies with different output lengths (x-axis) and input lengths (y-axis). This representation reveals a consistent non-linear pattern: the highest efficiency occurs at $n_{\text{in}} = 64$ and $n_{\text{out}} \in [128,256]$, whereas the lowest efficiency is observed for shortest output lengths $n_{\text{out}} = 64$ combined with the longest input length $n_{\text{in}} = 4096$.  These results suggest the existence of “sweet spots” where LLMs achieve the best trade-off between energy consumption and tokens generated.

Quantitatively, comparing for each model the worst efficient tokens combination, always located at the larger input and shorter output, with the highest energy efficiency configuration case, we observe on average a $33.41\times$ ($\pm 4.90$) increase in efficiency across models. 
In Appendix~\ref{subsec:appendix-parameters} (see \autoref{tab:fullpage-regression}), we report the $\theta$ parameters for our \textit{SweetSpot} model, along with the experimental and the estimated peak efficiency configurations, calculated using \autoref{eq:sweet-spots}.
\subsubsection{\textbf{Prediction of Peak Energy Efficiency Spots}}
\label{subsubsec:analysis}
To estimate the Energy-per-Token $E_{\text{tok}}$ for each combination of $n_{\text{in}}$ and $n_{\text{out}}$, we fitted the state-of-the-art Baseline 1-4 models and our proposed analytical model \textit{SweetSpot} described in \autoref{tab:model_set} with the results of the experiments of each LLM separately, using a non-linear least square method. In \autoref{tab:mape}, we report the Mean Absolute Percentage Error (MAPE) for each model fitted on the relative LLM.

We can observe that the proposed \textit{SweetSpot} model consistently achieves higher accuracy than the baselines for all the tested LLMs. On average, the Baselines 1 and 2 \cite{niu2025eeorexaustive} achieve a MAPE of 130.45\% and 130.49\%, respectively. Baseline 3 \cite{wilhelm2025beyondtestime} improves the MAPE on average to 106.25\%, and Baseline 4 \cite{wilkins2024offlineenergyoptimalllmserving} further reduces it to 31.40\%. 
This suggests that adding both $n_{\text{in}}$ and $n_{\text{out}}$ as fitting parameters improves the energy efficiency estimation of LLM inference w.r.t. considering only $n_{\text{out}}$. Our proposed model improves the MAPE over Baseline 4 significantly, reducing it to 2.27\% $(\pm0.61)$ and 1.79\% $(\pm0.61)$ in its FLOPs-only and complete versions respectively. This underlines that the FLOPs-only \textit{SweetSpot} model derived from computational complexity estimates captures better the dependency of LLM inference energy efficiency w.r.t. SoA, and that this dependency is non-linear with both quadratic and mixed terms dependencies. Furthermore, the memory extended \textit{SweetSpot} model consistently reduces the MAPE over the FLOPs-only one with an average reduction of 18.6\%, underlining that memory-access operations are as important as FLOPs and the sixth term $\frac{\theta_5}{n_{\text{out}}}$ in the analytical model is crucial.

In particular, the quadratic term $n_{\text{in}}^2$ can be interpreted as the initial cost incurred during the prefill phase. This cost dominates when the input sequence is relatively long, creating a high starting point in the per-token energy curve. As output tokens are generated, the cost reaches an optimal point where the amortization of input costs balances the linear per-token decoding costs. Beyond this optimum, the cost grows linearly with the number of output tokens, reflecting the sequential nature of decoding. 

Finally, by examining the derivative of the per-token energy with respect to $n_{\text{out}}$, we can identify the points of maximal efficiency observed in the experiments. The derivative analysis confirms the presence of an optimal number of output tokens that minimizes the per-token energy, matching the peak efficiency points observed in the empirical data.
\begin{table}
\footnotesize 
\setlength{\tabcolsep}{2pt} 
\caption{Mean Absolute Percentage Error (MAPE) achieved by each analytical and baseline model across the LLMs of the test set.}
\centering
\begin{tabularx}{\columnwidth}{l *{6}{>{\centering\arraybackslash}X}}
\toprule
\textbf{Model} &
\textbf{\scriptsize Baseline 1} &
\textbf{\scriptsize Baseline 2} &
\textbf{\scriptsize Baseline 3} &
\textbf{\scriptsize Baseline 4} &
\textbf{\scriptsize SweetSpot (FLOPs-only)} &
\textbf{\scriptsize SweetSpot} \\
\midrule
\multicolumn{7}{l}{\textbf{XS models}} \\
Llama 3.2 (1B)     & 102.33 & 99.45  & 93.58  & 17.84 & 2.95 & 1.56 \\
OPT (1.3B)         & 139.77 & 140.98 & 103.60 & 42.44 & 1.29 & 0.97 \\
Qwen 2 (1.5B)      & 110.71 & 100.90 & 103.60 & 14.28 & 2.62 & 1.94 \\

\midrule

\multicolumn{7}{l}{\textbf{S models}} \\
Gemma 2 (2B)       & 111.33 & 116.29 & 93.01  & 25.76 & 2.17 & 1.66 \\
OPT (2.7B)         & 153,29 & 153.59 & 111.64 & 47.14 & 1.47 & 0.91 \\
Llama 3.2 (3B)     & 117.54 & 121.55 & 99.77  & 24.19 & 2.46 & 2.12 \\
Granite (3B)       & 144.28 & 148.05 & 107.43 & 42.44 & 1.85 & 1.08 \\

\midrule

\multicolumn{7}{l}{\textbf{M models}} \\
OPT (6.7B)         & 151.27 & 155.47 & 110.08 & 46.49 & 1.82 & 1.93 \\
Qwen 2 (7B)        & 132.26 & 113.34 & 126.04 & 15.26 & 3.24 & 2.88 \\
Falcon-RW (7B)   & 150.80 & 155.75 & 109.90 & 47.30 & 1.48 & 1.35 \\
Granite (8B)       & 127.92 & 128.22 & 111.45 & 24.32 & 2.62 & 2.24 \\
Llama 3.1 (8B)     & 123.00 & 123.41 & 107.60 & 23.59 & 3.01 & 2.84 \\
Gemma 2 (9B)       & 131.38 & 139.41 & 103.54 & 37.17 & 2.49 & 1.81 \\
\midrule
\textbf{Average}   & \textbf{130.45} & \textbf{130.49} & \textbf{106.25} & \textbf{31.40} & \textbf{2.27} & \textbf{1.79} \\
\midrule
\textbf{Std Dev}   & \textbf{16.23} & \textbf{19.14} & \textbf{8.22} & \textbf{12.21} & \textbf{0.61} & \textbf{0.61} \\
\bottomrule
\end{tabularx}
\label{tab:mape}
\end{table}
\section{Limitations and Future work}
\label{sec:limitations_future_works}
Although the \textit{SweetSpot} model provides meaningful insights into LLM inference energy efficiency, its generality is limited by assumptions on model scale, hardware platform, inference framework, and batching configuration. This section discusses these limitations and outlines directions for future work.
\subsection{Experimental Setting}
The current study focuses on not very large models, which may not fully capture the performance characteristics of larger LLMs with substantially different parameter counts and memory footprints. Second, all experiments were conducted on a single hardware platform, limiting the generalizability of the results to heterogeneous systems with different compute, memory, and interconnect properties. Finally, we relied on a single inference framework, which constrains the scope of our conclusions, as alternative inference solutions may employ different optimization strategies and execution models affecting performance.

Future work will address these limitations by extending the experimental study to a broader range of model sizes, including large-scale and emerging LLM architectures, and by evaluating performance across multiple hardware platforms. In addition, incorporating multiple inference frameworks will allow a more systematic comparison of software-level optimizations and request scheduling strategies. Developing such a generalized performance model would provide a more comprehensive framework for predicting LLM inference efficiency and for guiding optimization strategies across a wider range of models and deployment scenarios.
\subsection{Batch Size Impact}
\label{subsec:batch-size_req}
\begin{figure}[t]
    \centering
    \subfloat[$N_{\text{req}}=128$\label{fig:batch_size_128req}]{
        \includegraphics[width=0.70\columnwidth]{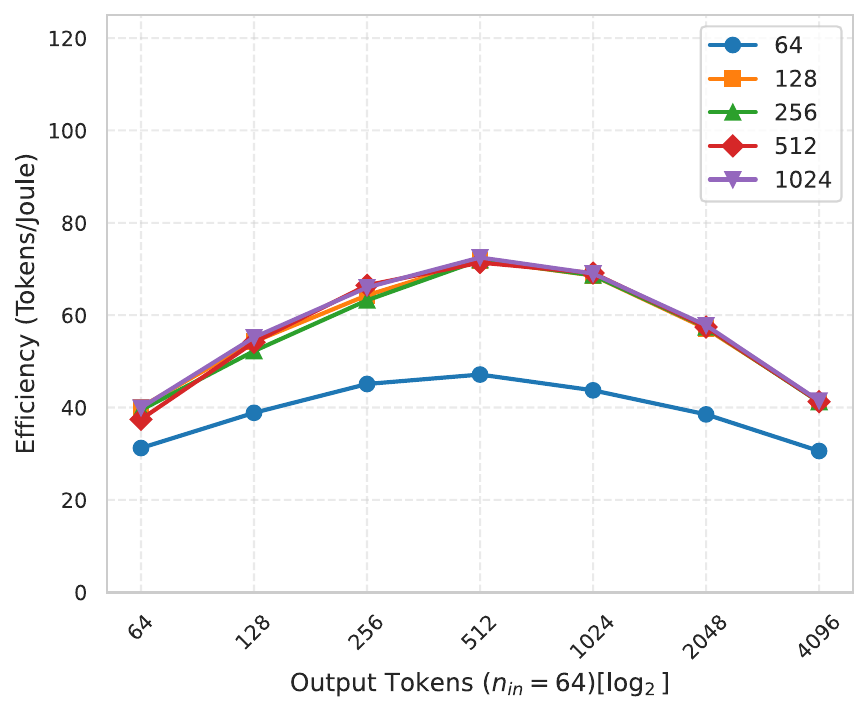}
    }
    \vspace{0.5em}    \subfloat[$N_{\text{req}}=1000$\label{fig:batch_size_1000req}]{
        \includegraphics[width=0.70\columnwidth]{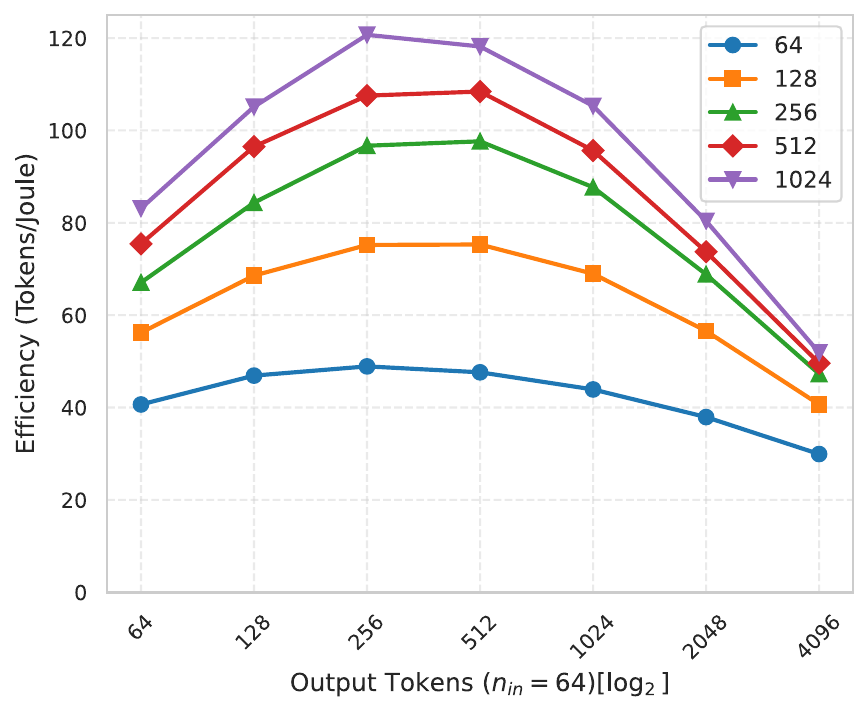}
    }

    \caption{Energy efficiency of Llama 3.2 1B varying maximum batch size considering different numbers of requests.}
    \Description{Two plots showing the energy efficiency of Llama 3.2 1B as maximum batch size varies, for $N_{\text{req}}=128$ and $N_{\text{req}}=1000$.}
    \label{fig:batch_size}
\end{figure}

The proposed \textit{SweetSpot} analytical model does not explicitly account for the batch size, instead assumes a fixed maximum batch size, which is set to 1024 in all evaluations. As shown in \autoref{fig:heatmaps_requests}, the number of concurrent requests has a strong impact on energy efficiency, particularly at peak values, as the engine can process a larger number of requests simultaneously. To analyze the interaction between the batch size and the number of requests, in \autoref{fig:batch_size} we report the energy efficiency of Llama 3.2 1B (the most efficient LLM in the test set) while varying the maximum batch size configuration (\texttt{\seqsplit{max\_batch\_size}}) across a range of values: 64, 128, 256, 512 and 1024. 
We tested these configurations for $N_{\text{req}}=128$ and $N_{\text{req}}=1000$, with the input sequence length fixed to $n_{\text{in}}=64$, where the maximum peak of energy efficiency occurs. 

In Figure  \ref{fig:batch_size_128req}, we observe that setting a maximum batch size larger than $N_{\text{req}}=128$ does not provide any efficiency benefit. Conversely, a smaller batch size like 64 leads to a significant reduction in efficiency, as this prevents the engine from forming larger batches, resulting in under utilization of computational resources. 

When considering a higher number of requests ($N_{\text{req}}=1000$), as illustrated in Figure \ref{fig:batch_size_1000req}, the maximum batch size has a strong effect on overall efficiency. In particular, the peak efficiency decreases from 120.71 with a batch size of 1024 to 48.90 when the batch size is reduced to 64. 
However, the impact of the batch size diminishes for longer output sequences. For instance, when $n_{\text{out}}=4096$, the efficiency gap between batch sizes of 256 and 1024 narrows to 47.27 and 51.91, respectively. This reduced sensitivity to batch size is attributable to the decode phase becoming dominant in requests with short inputs and long outputs, causing the engine to be constrained primarily by memory bandwidth rather than by the batch dimension.

In future work, we will extend our analysis of how maximum batch size affects LLM energy efficiency and explicitly incorporate this parameter into the proposed analytical model.
\section{Conclusions}
\label{sec:conclusions}
This paper presented a comprehensive study of the energy efficiency of LLM inference evaluated on NVIDIA H100 GPUs with TensorRT-LLM. We showed that energy consumption during inference can be explained by decomposing FLOPs and memory-access contributions into prefill and decode phases, and by normalizing them on a per-token basis. The resulting analytical model, \textit{SweetSpot}, highlights the quadratic overhead of input processing, the amortization effect of output length, and the linear cost of sequential decoding. Our experiments confirm that \textit{SweetSpot} aligns closely with real-world energy measurements. Specifically, we identified efficiency regimes where energy-per-token reaches a minimum, and demonstrated that these correspond to the balance point where input costs are sufficiently amortized without incurring excessive decoding overhead. This explains why efficiency peaks at short-to-moderate input lengths and medium outputs, while dropping significantly for long prompts or short generations.

Beyond theoretical insight, our findings have practical implications for LLM deployment. The analytical structure of inference costs allow to foresee energy efficiency "sweet-spots" configurations and to adapt prompt or generation length to reduce energy consumption. This perspective transforms energy efficiency from an empirical concept into a predictable design parameter. Building a generalized compute–memory efficiency model will further broaden the applicability, supporting sustainable optimization across different architectures and deployment contexts. In summary, our study represents an intermediate step between empirical benchmarking and theoretical modeling of LLM inference costs, providing a model that not only explains current efficiency regimes but also guides more energy-aware deployment strategies for large-scale generative AI.
\begin{acks}
The research has been partially funded by the EU Pilot for exascale EuroHPC EUPEX (g.a. 101033975), EuroHPC JU SEANERGYS (g.a. 101177590), DARE (g.a. 101143421) and Spoke “FutureHPC \& BigData” of the ICSC-Centro Nazionale di Ricerca in “High Performance Computing, Big Data \& Quantum Computing”, funded by the EU - NextGenerationEU projects.
\end{acks}
\bibliographystyle{ACM-Reference-Format}
\balance
\bibliography{ref}
\appendix
\begin{table*}
\centering
\caption{Aggregated statistical significance across LLM datasets.}
\begin{tabular}{lcccc}
\toprule
Parameter 
& $p < 0.001$ (***) 
& $p < 0.01$ (**) 
& $p < 0.05$ (*) 
& $p \geq 0.05$ (n.s.) \\
\midrule
$\theta_0$ & 13/13 (100\%) & 0/13 (0\%) & 0/13 (0\%) & 0/13 (0\%) \\
$\theta_1$ & 13/13 (100\%) & 0/13 (0\%) & 0/13 (0\%) & 0/13 (0\%) \\
$\theta_2$ & 13/13 (100\%) & 0/13 (0\%) & 0/13 (0\%) & 0/13 (0\%) \\
$\theta_3$ & 13/13 (100\%) & 0/13 (0\%) & 0/13 (0\%) & 0/13 (0\%) \\
$\theta_4$ & 13/13 (100\%) & 0/13 (0\%) & 0/13 (0\%) & 0/13 (0\%) \\
$\theta_5$ & 8/13 (62\%) & 1/13 (8\%) & 3/13 (23\%) & 1/13 (8\%) \\
\bottomrule
\end{tabular}
\label{tab:parameter_significance}
\end{table*}

\begin{table*}
\small
\caption{Theta values and Peak Energy Efficiency Configuration experimented and predicted by our SweetSpot model.}
\centering

\begin{tabularx}{\textwidth}{l *{8}{>{\centering\arraybackslash}X}}
\toprule

\textbf{Model} &
$\theta_0$ &
$\theta_1$ &
$\theta_2$ &
$\theta_3$ &
$\theta_4$ &
$\theta_5$ &
Peak $E_{\text{eff}}$ Config&
Peak $E_{\text{eff}}$ Config (Predicted) \\

\midrule
\multicolumn{9}{l}{\textbf{XS models}} \\
Llama 3.2 (1B)     & 5.005153e-03 & 1.079941e-07 & 6.825240e-06 & 2.611042e-03 & 3.852659e-06 & 5.406443e-01 & 64/256 & 64/429\\
OPT (1.3B)         & 5.969229e-03 & 1.666292e-07 & 4.388860e-05 & 2.915735e-03 & 2.342971e-05 & 3.187084e-01 & 64/64 & 64/147\\
Qwen 2 (1.5B)      & 6.528509e-03 & 9.147483e-08 & 6.199954e-06 & 3.700520e-03 & 3.288754e-06 & 3.805096e-01 & 64/512 & 64/433\\

\midrule
\multicolumn{9}{l}{\textbf{S models}} \\
Gemma 2 (2B)       & 9.468873e-03 & 1.223718e-07 & 2.486662e-05 & 5.371303e-03 & 1.372551e-05 & 5.856132e-01 & 64/256 & 64/260 \\
OPT (2.7B)         & 6.541956e-03 & 2.855932e-07 & 9.197051e-05 & 5.909706e-03 & 5.038964e-05 & 8.566784e-01 & 64/128 & 64/157 \\
Llama 3.2 (3B)     & 9.259155e-03 & 1.842970e-07 & 2.709305e-05 & 6.842019e-03 & 1.461407e-05 & 8.942031e-01 & 64/256 & 64/302 \\
Granite (3B)       & 9.300161e-03 & 4.210179e-07 & 9.340093e-05 & 7.914036e-03 & 5.199433e-05 & 1.170166e+00 & 64/128 & 64/180 \\

\midrule
\multicolumn{9}{l}{\textbf{M models}} \\
OPT (6.7B)         & 1.300082e-02 & 4.308046e-07 & 1.440099e-04 & 1.379572e-02 & 8.632801e-05 & 1.016691e+00 & 64/128 & 64/148 \\
Qwen 2 (7B)        & 1.857219e-02 & 2.598309e-07 & 1.531846e-05 & 1.465957e-02 & 9.735924e-06 & 8.719728e-01 & 64/512 & 64/431 \\
Falcon-RW (7.5B)   & 1.871769e-02 & 9.200243e-07 & 1.652377e-04 & 1.673305e-02 & 1.052294e-04 & 7.410144e-01 & 64/128 & 64/131 \\
Granite (8B)       & 2.073270e-02 & 3.926489e-07 & 4.029094e-05 & 1.744363e-02 & 2.512785e-05 & 8.540071e-01 & 64/256 & 64/280 \\
Llama 3.1 (8B)     & 1.970934e-02 & 3.329326e-07 & 3.583505e-05 & 1.551942e-02 & 2.249613e-05 & 8.939092e-01 & 64/256 & 64/290 \\
Gemma 2 (9B)       & 1.749322e-02 & 5.981696e-07 & 1.110917e-04 & 1.901597e-02 & 7.408055e-05 & 2.558409e+00 & 64/64 & 64/226 \\

\bottomrule
\end{tabularx}

\label{tab:fullpage-regression}
\end{table*}

\section{Appendix}
This appendix provides additional quantitative details supporting the modeling and statistical analysis presented in the main body of the paper. In particular, it reports the complete set of regression parameter significance statistics and values for our proposed \textit{SweetSpot} analytical model across all evaluated LLMs.
\subsection{Parameter Significance}
Table~\ref{tab:parameter_significance} summarizes the aggregated statistical significance of each analytical model parameter across all evaluated LLMs. By reporting the distribution of $p$-values for each parameter, this table highlights which components of the model consistently contribute to explaining performance variation and which parameters exhibit weaker or model-dependent influence. Together, these tables complement the main results by offering a detailed view of the model fitting behavior and its statistical robustness across different architectures and scales.
In the table, the number of asterisks quantify the significance of the parameter, with (***) meaning high significance, (**) moderate significance, (*) weaker significance and (n.s.) not significant. 
\subsection{Parameter Values}
\label{subsec:appendix-parameters}
Table~\ref{tab:fullpage-regression} lists the fitted values of the model parameters $\theta_0$--$\theta_5$ for each LLM, grouped by scale. In addition, it reports both the empirically observed  peak energy-efficiency configurations $(n_{\text{in}},n_{\text{out}})$ and the corresponding values estimated by our \textit{SweetSpot} model, enabling a direct comparison between measured and predicted optimal configurations. These results provide transparency into the parameterization of the model and allow reproducibility of the \textit{SweetSpot} prediction process.
\end{document}